\documentclass{article} 
\usepackage{iclr2026_conference,times}

\usepackage{amsmath,amsfonts,bm}

\def\eqref#1{equation~\ref{#1}}

\def\1{\bm{1}}

\DeclareMathAlphabet{\mathsfit}{\encodingdefault}{\sfdefault}{m}{sl}
\SetMathAlphabet{\mathsfit}{bold}{\encodingdefault}{\sfdefault}{bx}{n}

\usepackage{pifont}
\usepackage{ulem}
\usepackage{booktabs}
\usepackage{hyperref}
\usepackage{url}
\usepackage [dvipsnames]{xcolor}
\definecolor{c3}{HTML}{fe793d}
\usepackage{subcaption}
\usepackage{dsfont, mathtools, mathrsfs}
\usepackage{comment}
\usepackage{bm} 
\usepackage{graphicx}
\usepackage{subcaption}
\usepackage{amsmath, amsthm, amssymb, amsfonts}

\theoremstyle{definition}

\theoremstyle{remark}
\usepackage{algorithm}
\usepackage{algorithmic}
\usepackage{multirow}
\usepackage{bm}
\usepackage[flushleft]{threeparttable}
\usepackage{bm}
\usepackage{amsthm}
\usepackage{amsmath}
\usepackage{amssymb}
\usepackage{algorithm}
\usepackage{algorithmic}
\usepackage{cite}
\usepackage{multirow}
\usepackage{enumitem}
\usepackage{stfloats}
\usepackage{booktabs,caption}
\usepackage{lipsum}
\usepackage[flushleft]{threeparttable}
\usepackage{graphicx}
\usepackage{xcolor}
\usepackage{lipsum}
\usepackage{footmisc}
\usepackage{booktabs}   
\usepackage{multirow}   
\usepackage{makecell}   
\usepackage{array}      
\usepackage[utf8]{inputenc}
\usepackage{amsmath}
\usepackage{amssymb}
\usepackage{graphicx}
\usepackage{booktabs} 
\usepackage{caption} 
\usepackage{hhline}
\usepackage{comment}

\title{Otters: An Energy-Efficient Spiking \\ Transformer via Optical Time-to-First-Spike Encoding }

\author{Zhanglu Yan$^{1,*, \dag }$, Jiayi Mao$^{2,*}$, Qianhui Liu$^1$, Fanfan Li$^{2}$,  Gang Pan$^4$, Tao Luo$^3$, \\ \textbf{Bowen Zhu$^2$, Weng-Fai Wong$^1$} \\
Department of Computer Science\\
$^1$National University of Singapore, $^2$Westlake University, $^3$A*STAR, $^4$Zhejiang University \\
}
\iclrfinalcopy 
\begin{document}

\maketitle

\begin{abstract}


Spiking neural networks (SNNs) promise high energy efficiency, particularly with time-to-first-spike (TTFS) encoding, which maximizes sparsity by emitting at most one spike per neuron. However, such energy advantage is often unrealized because inference requires evaluating a temporal decay function and subsequent multiplication with the synaptic weights. 
This paper challenges this costly approach by repurposing a physical hardware `bug', namely, the natural signal decay in optoelectronic devices, as the core computation of TTFS. We fabricated a custom indium oxide optoelectronic synapse, showing how its natural physical decay directly implements the required temporal function. By treating the device's analog output as the fused product of the synaptic weight and temporal decay, optoelectronic synaptic TTFS (named Otters) eliminates these expensive digital operations. 
To use the Otters paradigm in complex architectures like the transformer, which are challenging to train directly due to the sparsity issue, we introduce a novel quantized neural network-to-SNN conversion algorithm. 
This complete hardware-software co-design enables our model to achieve state-of-the-art accuracy across seven GLUE benchmark datasets and demonstrates a 1.77$\times$ improvement in energy efficiency over previous leading SNNs, based on a comprehensive analysis of compute, data movement, and memory access costs using energy measurements from a commercial 22nm process. 
Our work thus establishes a new paradigm for energy-efficient SNNs, translating fundamental device physics directly into powerful computational primitives. All codes and data are open source\footnote{https://anonymous.4open.science/r/ICLR26Otters-26F1/README.md\\ Correspondence to Zhanglu Yan. Email: \{zlyan,qhliu, wongwf\}@nus.edu.sg, \{maojiayi,lifanfan, zhubowen\} @westlake.edu.cn, gpan@zju.edu.cn,luo$\_$tao@ihpc.a-star.edu.sg.}.

\end{abstract}

\section{Introduction}

Large language models (LLMs) have demonstrated remarkable capabilities, yet their immense computational and energy costs hinder their deployment in resource-constrained environments such as edge devices~\citep{lin2023pushing, jegham2025hungry}. This critical challenge has spurred research on more efficient, brain-inspired architectures, with {\em spiking neural networks} (SNNs) emerging as a promising candidate~\citep{tangsorbet,xingspikelm}. SNNs are known for their potential energy efficiency, which stems from sparse, event-driven computations that use addition instead of expensive multiplications. However, realizing this efficiency in practice is complex and depends heavily on the encoding scheme used~\citep{yan2024reconsidering}.
In conventional rate-coded SNNs, information is encoded in the number of spikes in a fixed time window. This approach necessitates multiple memory accesses for weights and frequent data movement on each spikes, which may negate the benefits of the sparse computation. Temporal coding schemes, specifically,  {\em time-to-first-spike} (TTFS), offer a potentially more efficient alternative. By encoding information in the precise timing of a single spike, a TTFS-SNN neuron fires at most once per activation cycle. This maximizes sparsity, dramatically reducing spike count and the associated data movement costs, making TTFS a theoretically optimal encoding for energy efficiency~\citep{yu2023ttfs,zhaottfsformer}.

\begin{figure}[htbp]
    \centering
    \includegraphics[width=0.9\linewidth]{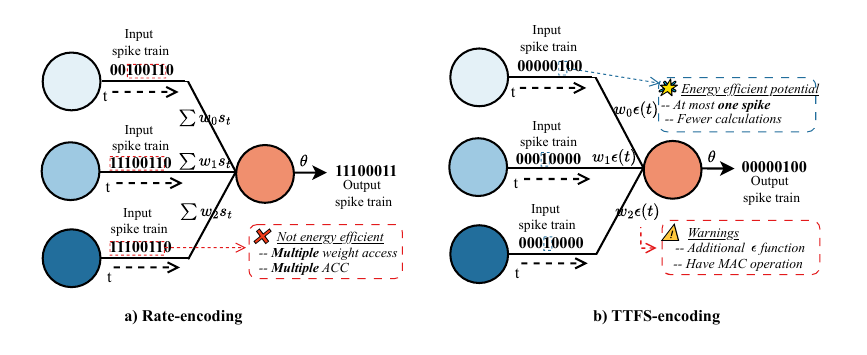}
    \caption{Rate encoding vs TTFS encoding }
    \label{fig:mac_acc}
\end{figure}

However, there are hidden costs behind this theoretical efficiency. In TTFS, information is encoded by mapping a spike's timing to its importance; typically, the earlier a spike arrives, the larger the numerical value it represents. To implement this principle, the network usually need to perform an extra computational step to convert the raw arrival time of each spike into a corresponding value. This is done using a decay function (e.g., $\epsilon(t) = e^{-t} $ or $T-t$) whose output is then multiplied by the synaptic weight ($w\cdot\epsilon(t)$)~\citep{wei2023temporal,che2024ettfs}. This conversion process requires energy to calculate the decay function itself, and it re-introduces the multiplication operations that SNNs are designed to avoid. This practical drawback negates the energy savings from sparsity and thus raising a critical question: how can we benefit from the sparsity of TTFS but avoid the costly computation?  Recognizing that the costly term $w\cdot\epsilon(t)$ is a value that predictably decreases over time, our answer lies not in optimizing the digital computation but finding a physical analog to simulate this process. This approach leads us to optoelectronic synapses, which are attractive for their fJ-level energy consumption and high resistance to electromagnetic interference~\citep{li2024artificial}. Notably, while this field has traditionally focused on suppressing their natural signal decay (volatility) to create stable memories~\citep{alqahtani2025light}, we embrace this decay. We recognize it not as a bug, but as the exactly physical implementation of the temporal decay function that TTFS requires. 
To implement this principle, we fabricated a custom $In_2O_3$  optoelectronic synapse. Our method, Otters, uses the natural decay of this device's optical signal to perform the required computation. 
This approach fundamentally integrates storage and computation into a single physical step, solving the overhead problem of traditional TTFS.

While our Otters hardware solves the computational overhead of TTFS, a second major barrier remains: the inherent difficulty of training such networks, especially for complex architectures like the Transformer. Directly training SNNs is challenging. In event-driven learning, error backpropagation depends on spike timing; if a neuron fires too sparsely or not at all, it fails to learn. This “over-sparsity" problem can severely limit the model's performance and even lead to training failure~\citep{wei2023temporal}.
To sidestep this issue, we employ a quantized neural network(QNN)-to-SNN conversion methodology. We first train a QNN and then convert its weights to our Otters SNN, avoiding the pitfalls of direct training in the spiking domain. To further maximize efficiency, we use knowledge distillation to train a highly compressed model with 1-bit weights and 1-bit key/value (KV) projections. Our complete hardware-software co-design, combining the Otters synapse with our QNN-to-SNN conversion pipeline, establishes a new state of the art for spiking language models. Evaluated on the GLUE benchmark, Otters achieves average accuracy of 3\% higher than previous leading SNNs while demonstrating a 1.77-3.04$\times$ improvement in energy efficiency compared to baselines like Sorbet and SpikingLM~\citep{tangsorbet,xingspikelm}. Notably, this energy efficiency gain is validated by a rigorous and comprehensive analysis that moves beyond the simplistic metrics common in prior SNN research. While previous work often only counted compute operations (e.g., additions vs. multiplications), our analysis provides a more realistic estimate. It is grounded in measurements from a commercial 22nm process and provides a full accounting of compute, data movement, and memory access costs, making our efficiency claims robust. 

We also investigate the Otters paradigm's sensitivity to the hardware noise inherent in analog devices. Our initial analysis shows that, on the SST-2 benchmark, the baseline model's performance begins to degrade with around 5\% variation in key physical parameters. To improve the robustness, we propose Hardware-Aware Training, a method where we introduce different levels of simulated Gaussian noise during the training process. This approach enhances the model's resilience, enabling it to maintain robust performance in noisy conditions, demonstrating a practical path toward robust real-world deployment.

 \section{Preliminary}

 \subsection{Optoelectronic synapse}
 
 An optoelectronic synapse is a neuromorphic device that emulates biological synaptic functions by using optical signals to modulate its electrical conductance. These devices are renowned for their potential for extreme energy efficiency, broader bandwidth and faster signal transmission in neuromorphic computing, which are key advantages over purely electronic counterparts~\citep{xie2024emerging,wang2023advanced}. Recent studies have reported energy consumption reaching the femtojoule (fJ)/spike level, comparable to biological synapses and substantially lower than conventional CMOS neuron devices~\citep{shi2022fully, wang2024monolithic}. Among various implementations, oxide thin-film transistors (TFT) are regarded as viable candidates for optoelectronic synapses due to their low leakage current and capability for large-area, flexible fabrication. Solution-based fabrication further offers the advantages of low cost, simplified processing, and facile compositional control. Previous reports have shown that solution-processed devices exhibit uniform performance, operational stability, and low energy consumption~\citep{li2025double}.  Building upon these advances, this work employs the mature and reliable oxide-TFT platform to develop the Otters spiking neuron.



 \subsection{Time to first spike SNN}
In contrast to rate-based encoding, which uses the frequency of spikes to represent information, TTFS encoding leverages the precise timing of a single spike. The core principle is that a stronger input stimulus causes a neuron's membrane potential to rise faster, reaching its firing threshold sooner. Thus, the information is encoded in the arrival time of the first—and only—spike within a given time window, $T$. This approach maximizes temporal sparsity and is highly efficient, as each neuron fires at most once~\citep{che2024efficiently}.

The operation of a standard TTFS neuron involves two phases. First, the neuron integrates incoming spikes, updating its membrane potential $V_j^l(t)$. Second, it compares this potential to a firing threshold $\theta^l(t)$. A spike is generated at the first time step $t$ where the potential meets or exceeds the threshold:
\begin{equation}
s_j^l(t) = 
\begin{cases}
1, & \text{if } V_j^l(t) \ge \theta^l(t) \\
0, & \text{otherwise}
\end{cases}
\end{equation}

However, the asynchronous nature of SNNs, combined with the ``fire-as-early-as-possible'' objective of TTFS, can lead to another problem. If a presynaptic neuron fires after a postsynaptic neuron has already fired,  its spike becomes invalid for membrane potential accummalation. To solve this, we employ a Dynamic Firing Threshold (DFT) model that enforces a synchronous, layer-by-layer processing schedule~\citep{wei2023temporal}. The threshold for any neuron in layer $l$ is set to infinity outside of a designated time window, effectively ensuring that layer $l$ is only active from time $T \cdot l$ to $T \cdot (l+1)$:
\begin{equation}
\theta^l(t) = 
\begin{cases}
\theta^l_{\text{dynamic}}(t), & \text{if } T \cdot l \le t \le T \cdot (l+1) \\
+\infty, & \text{otherwise}
\end{cases}
\end{equation}
This scheduling guarantees that all spikes from a preceding layer are processed before the current layer can fire, thus preserving the valid causal relationship.

 \section{Methods}
 This section details the methodology behind Otters. We first describe the core of our model: the optoelectronic synapse and the neuron model that performs the TTFS computation (Section~\ref{sec:otters_neuron}). We then explain how these components are assembled into the complete optimized spiking Transformer architecture (Section~\ref{sec:network_structure}), and then, we outline the QNN-to-SNN conversion pipeline used to build the Otters model (Section~\ref{sec:qnn_to_snn}). Finally, we present the framework for the comprehensive energy analysis used to validate our model's efficiency (Section~\ref{sec:energy_analysis}).


 \begin{figure}[htbp]
    \centering
    \includegraphics[width=0.9\linewidth]{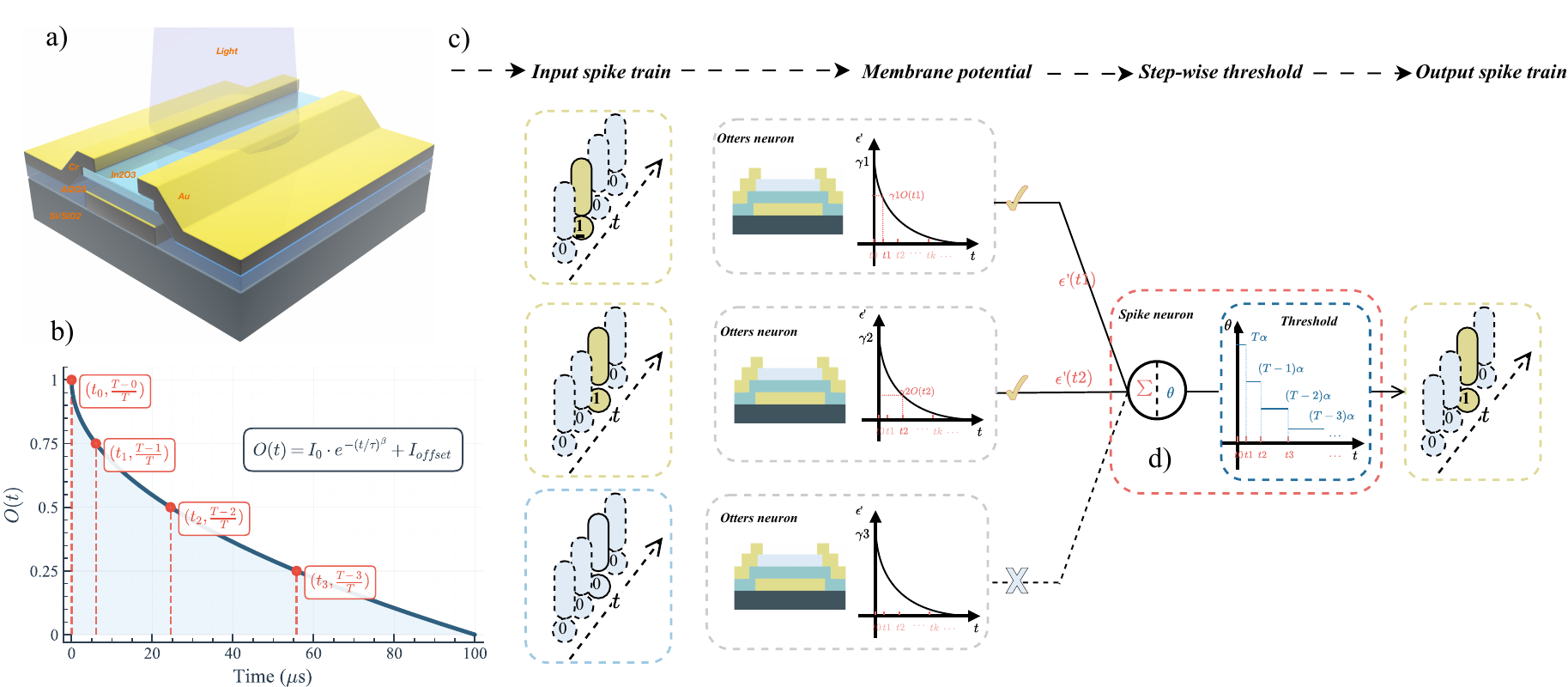}
\caption[Device and workflow]{Device and workflow: (a) the custom-fabricated In$_2$O$_3$ thin-film transistor (TFT); (b) measured decay curve of the device response; (c) Otters neuron workflow.}

    \label{fig:otters}
\end{figure}

\subsection{OTTERS SPIKING NEURON}
\label{sec:otters_neuron}
The core of our method is the Otters optoelectronic synapse, a hardware element that physically implements the time-modulated synaptic dynamics required for TTFS computation. Each synapse is composed of two main parts: a custom-fabricated Indium Oxide Thin-Film Transistor that provides a physical signal decay, and an analog-to-digital converter (ADC) that map and scale the analog signal to digital. Fabrication details for the device in Fig.~\ref{fig:otters}(a) are provided in Appendix~\ref{sec:Fabricated }. To ensure a deterministic response, the TFT is operated with a fixed light intensity, yielding a consistent non-linear decay curve, modeled by the function 
\(O(t) = I_0 \cdot e^{-\left(t/\tau\right)^{\beta}} + I_{\text{offset}}\)~\citep{li2024active,liang2022printable}
. We fit the model parameters by minimizing the sum of squared residuals using the differential evolution algorithm, yielding: $I_0=110.989$, $\tau=1.3425$, $\beta=0.495$, and $I_{\text{offset}}=-109.989$, showing in Figure~\ref{fig:otters}(b).  Thus, the TFT's current decay over time naturally forms the temporal component of the post-synaptic potential (PSP).

However, the physical non-linearity of the device presents a critical design challenge. For our QNN-to-SNN conversion to be lossless, the information encoded by a spike's timing must map to a set of uniformly spaced logical values. Specifically, a spike occurring at a physical time $t_k$ must represent the quantized value $(T-k)/T$. The device's non-linear decay, $O(t)$, means that the physical times $t_k$ at which the device output naturally equals these target values are themselves non-uniformly spaced. A naive approach using a constant threshold and uniform time sampling would therefore fail to establish the required functional equivalence. 

Therefore, our solution is to reconcile the non-linear device physics with the linear encoding requirement. 
Instead of implementing a complex, non-uniform clock, we engineer a dynamic, step-wise decreasing firing threshold, $\theta^l(t)$, while operating the system on a standard, uniform physical clock. This threshold is designed to change its value only at the pre-calculated time points $\{t_k\}$ which are derived from the inverse of the physical decay function. This design ensures that the firing condition where the membrane potential exceeding the threshold can only be met at one of these discrete moments $t_k$. The neuron fires at the first such time point where its accumulated potential is sufficient. Consequently, the output spike time $t_k$ can encode the intended quantized value $(T-k)/T$. Because this encoding scheme is applied consistently throughout the network, the output spike of one layer provides a correctly timed and valued input to the next, ensuring the integrity of information propagation across the entire architecture. The full mathematical formulation of this threshold will be defined in our conversion methodology in \text{Section~\ref{sec:qnn_to_snn}}. 

The ADC implements a \(\gamma_{ij}^{l}\) times scaling mapping from the physical decay to a digital post-synaptic potential (PSP). Thus, the full PSP $\epsilon^{\prime}$ generated by a presynaptic spike is therefore the direct analog output of the synapse at the time of arrival:
\begin{equation}
\label{eq:psp}
\epsilon^{\prime}(t) = \gamma_{ij}^{l} \cdot O(t)
\end{equation}
The membrane potential \(V_{j}^{l}(t)\) of neuron j accumulates these PSPs. At each discrete physical time step t, the potential is updated based on incoming spikes:
\begin{equation}
\label{eq:membrane_update}
V_{j}^{l}(t) = V_{j}^{l}(t-1) + \sum_{i \text{ s.t. } s_{i}^{l-1}(t)=1} \epsilon^{\prime}(t)
\end{equation}
A neuron fires when its membrane potential first meets or exceeds the dynamic threshold. This spike time, \(t_{\text{spike},j}^{l}\), corresponds to the first logical timestep k where the condition is met:
\begin{equation}
\label{eq:spike_time}
t_{\text{spike},j}^{l} = \min\{t_{k} | V_{j}^{l}(t_{k}) \ge \theta^{l}(t_{k})\}
\end{equation}
In adherence with the TTFS paradigm, the neuron is deactivated after firing to ensure at most one spike per inference cycle. Algorithm ~\ref{alg:otters_neuron} in appendix formally describes this forward pass.

\subsection{Network Structure }
\label{sec:network_structure}

\begin{figure}[h!]
  
    \begin{minipage}[b]{0.49\textwidth}
        \centering
        \includegraphics[width=0.9\textwidth]{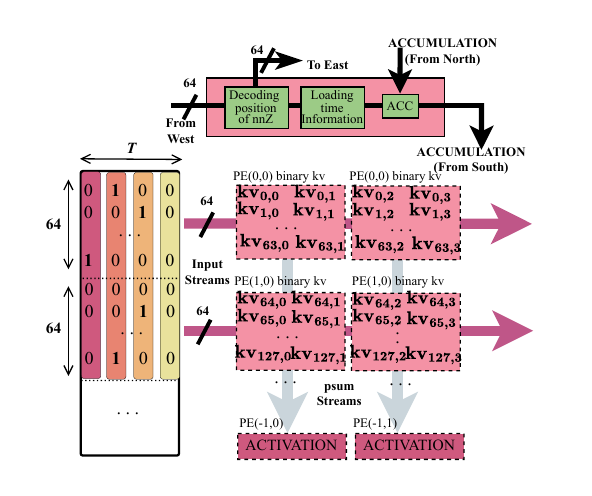}
        \caption{Scores calculations}
        \label{fig:scores_cal}
    \end{minipage}
    \hfill 
    \begin{minipage}[b]{0.49\textwidth}
        \centering
        \includegraphics[width=1.09\textwidth]{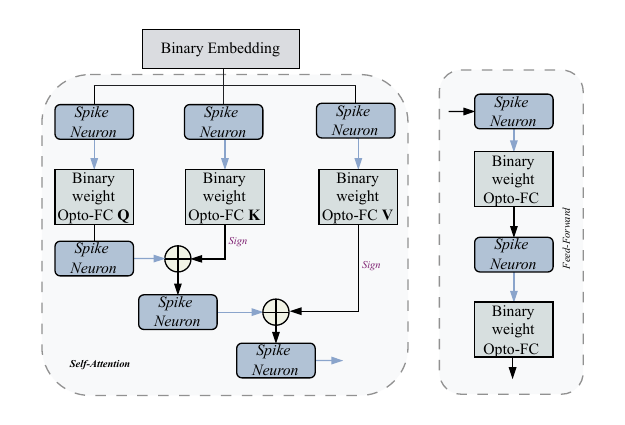}
        \caption{Otters-based transformer structure}
        \label{fig:main_structure}
    \end{minipage}
\end{figure}

A primary challenge in creating a spiking Transformer is the matrix multiplication required for self-attention score calculation (\(Q\cdot K^{T}\)). While some rate-coded SNNs can simplify this by treating one matrix as a binary spike train (turning multiplication into selective addition), this approach is incompatible with TTFS encoding, which requires decoding spikes into non-binary values.  
To overcome the problem, we quantize the key ($K$) and value ($V$) projections to a single bit, $\{+1,-1\}$. Consequently, the dot product with a TTFS-encoded query ($Q$) is computed using only selective, additions and subtractions. This allows us to eliminate the multiplication bottleneck while still benefiting from the high sparsity of TTFS. 
To implement this 1-bit attention mechanism efficiently, we designed a supporting dataflow architecture inspired by the Canon architecture~\citep{bai2025data}, as illustrated in Figure~\ref{fig:scores_cal}. During inference, the binary K (or V) vectors are pre-loaded into the local memory of a Processing Element (PE) array. The TTFS-encoded input stream (representing Q) is broadcast to the PEs. As shown, each PE computes a partial sum by accumulating its local K values only at the time steps corresponding to incoming spikes. These partial sums are then passed between PEs for final accumulation. This architecture minimizes data movement and leverages the spatio-temporal sparsity of the TTFS input for energy efficiency.

This self-attention module, along with feed-forward layers built from our Opto-FC and spiking neuron primitives, forms the complete Otters Transformer architecture shown in Figure~\ref{fig:main_structure}.

\subsection{QNN-to-SNN Conversion}
\label{sec:qnn_to_snn}

To overcome the challenges of direct SNN training, we employ a QNN-to-SNN conversion methodology. We first train a QNN and then map its learned parameters to an equivalent Otters SNN, ensuring the two networks are functionally identical. To formalize this relationship, we first define the specific QNN layer architecture that enables this equivalence. A compatible QNN layer computes its output \(x_{q,j}^{l}\) for neuron j as follows:
\begin{equation}
\label{eq:qnn_pre_activation}
a_{j}^{l}=\sum_{i}w_{ij}^{l}x_{q,i}^{l-1}+b_{j}^{l}
\end{equation}
\begin{equation}
\label{eq:qnn_activation}
x_{q,j}^{l}=Q(a_{j}^{l})=\alpha^{l}\cdot \text{Clip}(\lfloor\frac{a_{j}^{l}}{\alpha^{l}}\rfloor,0,2^{n}-1)
\end{equation}
where \(a_{j}^{l}\) is the pre-activation, \(w_{ij}^{l}\) are the weights, \(b_{j}^{l}\) is the bias, and \(\alpha^{l}\) is the quantization scaling factor for layer l. With this structure established, we can now state the proposition that governs the exact conversion from the trained QNN to the Otters SNN:

\textbf{Proposition 1.} \textit{An Otters SNN layer (as defined in Section~\ref{sec:otters_neuron}) is functionally equivalent to a trained n-bit QNN layer (as defined above) if its parameters are constructed as follows:}

\begin{enumerate}
    \item \textit{The number of discrete time steps in the SNN simulation window, T, is set to match the number of positive quantization levels of the n-bit QNN: \(T=2^{n}-1\).}
    \item \textit{The mapping from a logical time step \(k \in \{0, 1, \dots, T-1\}\) to a physical spike time \(t_k\) is defined such that the device's output at that instant, \(O(t_k)\), is linearly proportional to the remaining time in the window: \(O(t_{k})=\frac{T-k}{T}\).}
    \item \textit{The SNN's physical scaling factor for the synapse connecting neuron i in layer \(l-1\) to neuron j in layer l, \(\gamma_{ij}^{l}\), is set based on the corresponding QNN weight and the quantization scale of the previous layer: \(\gamma_{ij}^{l}=w_{ij}^{l}\cdot\alpha^{l-1}\cdot T\).} 
    \item \textit{The SNN's firing threshold for neuron j in layer l is a step-wise decreasing function of time, defined as: \(\theta^{l}(t)=\alpha^{l}\cdot(T-k)\), for \(t_{k} \le t < t_{k+1}\).}
\end{enumerate}

The proof proceeds in two steps. First, during integration, we show that the accumulated membrane potential \(V_{j}^{l}\) in an SNN neuron is numerically identical to the corresponding QNN neuron's pre-activation value \(a_{j}^{l}\). e binarize weights $w^l_{ij}$ to reduce energy and increase the reuse factor $\gamma^l_{ij}$. Second, during firing, we show that the engineered, time-dependent threshold \(\theta_{j}^{l}(t)\) compensates for the device's intrinsic non-linearity by permitting firing only at pre-calculated time points, \(t_k\), derived from the inverse of the physical decay function. This mechanism ensures the SNN neuron fires at a time \(t_{\text{spike},j}^{l}\) that precisely encodes the QNN's quantized output \(x_{q,j}^{l}\). A detailed proof is provided in the Appendix~\ref{sec: proof}.

\subsection{Energy Analysis}
\label{sec:energy_analysis}

To evaluate the efficiency of our approach, we formulate an analytical energy model. The total inference energy ($E$) is decomposed into three primary components: computation energy ($E_{\text{Compute}}$), data movement energy ($E_{\text{Data}}$), and analog energy ($E_{\text{Analog}}$). The computation energy, $E_{\text{Compute}}$, accounts for arithmetic operations of additions. The data movement energy, $E_{\text{Data}}$, encompasses the energy for transferring data, including both dynamic and static power consumption. The final component, $E_{\text{Analog}}$, includes the energy to power the TFT, the sampling energy, and ADC energy required to converted the analog signal to scaled digital value, collectively represented as \(E^{\text{Read}}_{\text{Analog}}\). Thus, we have \(
    E = E_{\text{Compute}} + E_{\text{Data}} +E_{\text{analog}}
\).
For our energy calculation, we consider a spatial dataflow architecture where information (e.g., spike packets) is communicated over a Network-on-Chip (NoC)~\citep{yan2024reconsidering}. This architecture is representative of modern specialized hardware such as neuromorphic chips like Loihi~\citep{lines2018loihi} and dataflow AI accelerators like Tenstorrent~\citep{tenstorrent} and Sambanova~\citep{prabhakar2022sambanova}. We consider the control logic energy to be negligible as our analysis focuses on specialized accelerator designs where such overhead is minimal~\citep{yan2024reconsidering}.

Thus, we model the energy consumption of our \text{Otters}-based linear projections and attention operations. We disregard the computational cost of certain operations, such as Softmax and Layer Normalization, as their contribution to the total compute is negligible compared to large-scale matrix multiplications. The energy to perform a linear projection, $E_{\text{Opto-FC}}$, is modeled by Equation~\ref{eq: opt-fc}. The calculation is performed over a batch of size $B$, a sequence of length $S$, and for $C_o$ output channels. The total energy is the sum of the following components per output element:

\begin{equation}
\label{eq: opt-fc}
\begin{split}
    E_{\text{Opto-FC}} = \underbrace{B \cdot S \cdot C_o}_{\text{Total Outputs}} \cdot \bigg( 
    &\underbrace{C_i \cdot T \cdot \big(s_r \cdot (E_{\text{ACC}} + E^{\text{Read}}_{\text{Analog}} + E^{\text{sparse}}_{\text{move}}) }_{\text{Spike Processing}}  \\
    & + E_{\text{leakage}}\big) + \underbrace{T \cdot (E_{\text{CMP}}+E^{\text{Read}}_{\text{threshold}})}_{\text{Thresholding}} + \underbrace{E^{\text{Write}}_{\text{binarykv}}}_{\text{K/V Write}} \bigg)
\end{split}
\end{equation}

\begin{itemize}
    \item Computation Energy: This includes the energy for sparse accumulations. The total number of active accumulate operations is the workload ($C_i$) scaled by the average number of spikes ($T \cdot s_r$), with each operation costing $E_{\text{ACC}}$. Additionally, each of the $C_o$ output neurons performs $T$ comparisons against its threshold, costing $E_{\text{CMP}}$ per comparison.

    \item Data Movement Energy: This is composed of dynamic and static costs. Dynamic energy ($E^{\text{sparse}}_{\text{move}}$) is consumed to move spike data and is proportional to the spike rate ($s_r$).
    Otters paradigm requires a dynamic, step-wise decreasing threshold, which introduces additional dynamic energy of reading operations $E^{\text{Read}}_{\text{threshold}}$ which is need for spiking neurons.
    Static energy ($E_{\text{leakage}}$) accounts for constant leakage power over the time window $T$. After the computation completes, an additional energy cost, $E^{\text{Write}}_{\text{binarykv}}$, is incurred to write the generated binary Key and Value vectors to SRAM, as described in Section~\ref{sec:network_structure}.

    \item Analog Energy: Each incoming spike initiates two analog operations: first, the optoelectronic synapse emits light, and second, the decay function is sampled at a specific time $t_k$. The energy for both the light emission and the sampling event is calculated by integrating the instantaneous power ($P = V \cdot I$) over the duration of each respective operation. For the readout circuitry, including the amplifier and the look-up-table, we adopt the energy values of a successive approximation register-assisted pipelined ADC~\citep{su20235}. The total energy consumed in this process is denoted $E^{\text{Read}}_{\text{Analog}}$ per spike.
    
\end{itemize}

The energy model for attention score calculation, $E_{\text{Opto-score}}$, is analogous to the linear projection, with two key differences which lies in the outer dimensions and an additional data movement cost for reading the binary Key (or Value) vector from SRAM to determine whether the sampled membrane potential should be added to or subtracted from the accumulator, showing in Appendix~\ref{sec: energy_qnn}.

\section{Results}
\label{sec:results}

In this section, we evaluate Otters on seven datasets from the GLUE benchmark. We compare its performance against both standard QNN and SNN baselines, using $\text{BERT}_{\text{base}}$ as the teacher model for knowledge distillation. We further provide a detailed analysis of the model's energy efficiency and robustness. All experiments were conducted on three NVIDIA A100 GPUs with a fixed 4-bit simulation window size recommended by Sorbet~\citep{tangsorbet} (In our setting, it is equal to timestep $T=15$). The training process of Otters is shown in Appendix~\ref{sec: training algo} and the description of dataset we adapted is shown in Appendix~\ref{sec: datasets}.

\subsection{GLUE Benchmark Performance}

As shown in Table~\ref{tbl:glue_performance_avg}, Otters achieves SOTA results among SNNs across all seven evaluated GLUE tasks, consistently outperforms larger and more complex SNN models like SpikingBERT and SpikeLM.
For example,  Otters surpasses existing SNNs and achieve an accuracy of \text{68.95\%} on RTE and 91.28\% on SST-2. The average accuracy for Otters is 83.22\%, which is 3.42\% and 2.98\% higher than Sorbet and SpikeLM, respectively.

\begin{table*}[h!]
\small
\centering
\caption{Performance comparison on the GLUE benchmark. All scores are accuracy, except for STS-B (Pearson correlation). “*” indicates that the model size was not reported in the original paper. \textbf{Bold} indicates the best performance among SNN models.}
\label{tbl:glue_performance_avg}
\resizebox{1.02\textwidth}{!}{%
\begin{tabular}{@{}lcccccccccc@{}}
\toprule
\textbf{Model} & \textbf{Size} & \textbf{QQP} & \textbf{MNLI-m} & \textbf{SST-2} & \textbf{QNLI} & \textbf{RTE} & \textbf{MRPC} & \textbf{STS-B} & \textbf{Average} \\
\midrule
$\text{BERT}_{\text{base}}$~\citep{devlin2019bert} & 418M & 91.3 & 84.7 & 93.3 & 91.7 & 72.6 & 88.2 & 89.4 & 87.31 \\
DistilBERT~\citep{sanh2019distilbert} & 207M & 88.5 & 82.2 & 91.3 & 89.2 & 59.9 & 87.5 & 86.9 & 83.64 \\
$\text{TinyBERT}_6$~\citep{jiao2020tinybert} & 207M & - & 84.6 & 93.1 & 90.4 & 70.0 & 87.3 & 83.7 & 84.85 \\
Q2BERT~\citep{zhang2020ternarybert}      & 43.0M & 67.0 & 47.2 & 80.6 & 61.3 & 52.7 & 68.4 & 4.4 & 54.51 \\
BiT~\citep{liu2022bit}     & 13.4M & 82.9 & 77.1 & 87.7 & 85.7 & 58.8 & 79.7 & 71.1 & 77.57 \\
SpikingFormer~\citep{zhou2023spikingformer}  & * & 83.8 & 67.8 & 82.7 & 74.6 & 58.8 & 74.0 & 72.3 & 73.43 \\
SpikingBERT~\citep{bal2024spikingbert}  & 50M & 86.8 & 78.1 & 88.2 & 85.2 & 66.1 & 79.2 & 82.2 & 80.83 \\
SpikeLM~\citep{xing2024spikellm}   & * & 87.9 & 76.0 & 86.5 & 84.9 & 65.3 & 78.7 & 84.3 & 80.51 \\
\midrule
1-bit SpikeLM ~\citep{xing2024spikellm} & * & 87.2 & 74.9 & 86.6 & 84.5 & 65.7 & 78.9 & 83.9 & 80.24 \\
1-bit Sorbet~\citep{tangsorbet} & 13.4M & 86.5 & 77.3 & 90.4 & 86.1 & 60.3 & 79.9 & 78.1 & 79.80 \\

\textbf{Otters (Ours)} & 13.4M & \textbf{87.67} & \textbf{78.50} & \textbf{91.28} & \textbf{86.42} & \textbf{68.95} & \textbf{84.56} & \textbf{85.19} & \textbf{83.22} \\
\bottomrule
\end{tabular}}
\end{table*}

\subsection{Energy Efficiency}
We analyzed the energy consumption of Otters on the SST-2 dataset, comparing it against full-precision and quantized BERT models we converted from, as well as SOTA SNNs. The 1-bit quantized BERT is the QNN Otters converted from which sharing the same structure and parameters.
As detailed in Table~\ref{tab:energy_consumption}, Otters is consumes only \text{4.06 mJ} per inference for one attention block. This represents a \text{41.36}$\times$ energy saving compared to the full $\text{BERT}_{\text{base}}$ model and a \text{2.72}$\times$ efficient compared to the 1-bit quantized BERT. Otters is also more efficient than previous SNNs, taking \text{3.04}$\times$ energy saving of Sorbet and \text{1.77}$\times$ of SpikingLM. 
The full energy equation, detailed measurements for all compared models, and an ablation of traditional TTFS vs Otters are provided in Appendix~\ref{sec: energy_qnn}. All energy figures include compute, data movement, and static components.

\begin{table}[h!]
\centering
\caption{Energy consumption analysis on the SST-2 dataset. Energy is reported for FC layers, QKV self-attention score calculation, and the total attention block (all FC and score calculation in Figure4) per inference. The Energy Ratio is Energy(Full BERT) / Energy(Model).}
\label{tab:energy_consumption}
\resizebox{0.8\textwidth}{!}{
\begin{tabular}{@{}lcccc@{}}
\toprule
\textbf{Model} & \textbf{FC (mJ)} & \textbf{QKV (mJ)} & \textbf{Total (mJ)} & \textbf{Energy Ratio} \\
\midrule
Full BERT~\citep{devlin2019bert} & 50.35 & 8.41 & 167.92 & 1.00x \\
1-bit Quantized BERT & 3.31 & 0.55 & 11.03 & 15.2x \\
Sorbet~\citep{tangsorbet} & 3.39 & 1.08 & 12.34 & 13.61x \\
SpikingBERT~\citep{bal2024spikingbert} & 6.37 & 2.05 & 23.22 & 7.23x \\
SpikingLM~\citep{xing2024spikellm} & 2.09 & 0.46 & 7.2 & 23.32x \\
\textbf{Otters (Ours)} & \textbf{1.14} & \textbf{0.33} & \textbf{4.06} & \textbf{41.36x} \\
\bottomrule
\end{tabular}}
\end{table}

\subsection{Effect of KV Cache Quantization}
To further optimize energy, we explored the impact of quantizing the Key and Value projections in the self-attention mechanism. Table~\ref{tab:kv_quantization} shows that reducing the KV precision from 4-bit to 1-bit (\text{Otters-1bitkv}) yields a \text{10\%} reduction in total energy consumption (from 4.49 mJ to 4.06 mJ). This energy saving comes at the cost of 0.23\% drop in accuracy on SST-2, demonstrating a highly favorable trade-off between efficiency and performance.

\begin{table}[hbt]
\centering
\caption{Impact of KV cache quantization on energy and accuracy on SST-2.}
\label{tab:kv_quantization}
\scalebox{0.8}{
\begin{tabular}{@{}l ccc c@{}}
\toprule
& \multicolumn{3}{c}{\textbf{Energy (mJ)}} & \\
\cmidrule(lr){2-4}
\textbf{Model} & \textbf{FC} & \textbf{QKV} & \textbf{Total} & \textbf{Accuracy (\%)} \\
\midrule
Otters-4bitkv & 1.14 & 0.53 & 4.49 & 91.51 \\
Otters-1bitkv & 1.14 & 0.33 & 4.06 & 91.28 \\
\bottomrule
\end{tabular}}
\end{table}

\subsection{Robustness Discussion to Hardware Parameter Variations}

The practical deployment of the Otters paradigm hinges on the assumption that the physical decay characteristics of all optoelectronic synapses are uniform. However, analog hardware is inevitably subject to device-to-device variability from the fabrication process, which is a significant challenge for large-scale integration~\citep{garg2022dynamic}. Such inherent hardware noise can corrupt the precise time-to-value mapping that underpins our conversion method. To evaluate robustness against hardware variability~\citep{fagbohungbe2021benchmarking,xuan2022low,su2024oxygen}, we injected zero-mean standard deviation Gaussian noise, which is commonly used to simulate and represent hardware noise, into the physical decay function, $O(t)$, and its parameters, $\tau$ and $\beta$. 
The injected noise is proportional to parameter magnitude: at level \(k\), each parameter \(p\) is scaled as \(p \leftarrow p\,(1 \pm k)\).
As shown in Table~\ref{tab:noise_ablation}, the baseline \text{Otters} model can tolerate about \(5\%\) total output \(O(t)\) difference  with minimal accuracy impact and remains robust. 



To keep improving the robustness, we propose \text{Hardware-Aware Training (HAT)}, a method that builds robustness by simulating hardware non-idealities during training. We introduce two variants, \text{HAT$^1$} and \text{HAT$^2$}, by injecting 10\% and 20\% Gaussian noise, respectively, into the QNN's activations (see Eq.~\ref{eq:qnn_activation}). As shown in Table~\ref{tab:noise_ablation}, both HAT settings improve noise resilience. The HAT$^2$ model maintains a stable accuracy of 80.8\% even under a 20\% noise level. HAT$^2$, trained with more noise, excels in high-noise regimes, whereas HAT$^1$ achieves higher accuracy in low-noise conditions while still substantially outperforming the baseline (e.g., a 11.5\% accuracy gain with 12\% noise in $O(t)$). Thus, HAT-trained models trade a minor drop in peak accuracy for a significant increase in resilience against hardware noise. This demonstrates that the model can be regularized to generalize across a range of hardware imperfections. Consequently, the HAT noise level can be tuned to the manufacturing tolerances of a specific hardware platform, ensuring reliable real-world performance and validating the Otters paradigm as a robust and practical approach.

\label{sec:robustness}
\begin{table*}[htbp]
\centering
\caption{Noise Robustness in Physical Decay Function. \textbf{Bold} indicates best performance per noise level within each component group. Results: mean $\pm$ std over 3 runs.}
\label{tab:noise_ablation}
\small
\resizebox{0.7\textwidth}{!}{%
\begin{tabular}{l|ccccc}
\toprule
\textbf{Method} & \multicolumn{5}{c}{\textbf{Gaussian Noise Level}} \\
\midrule
\multicolumn{6}{c}{\textbf{Full Function } $\mathbf{O(t)}$ \textbf{ Experiments}} \\
\cmidrule(lr){1-6}
$O(t)$& \textbf{0.04} & \textbf{0.08} & \textbf{0.12} & \textbf{0.16} & \textbf{0.20} \\
\midrule
Otters & \textbf{89.9} $\pm$ \textbf{0.8} & 86.1 $\pm$ 0.8 & 73.8 $\pm$ 0.9 & 58.0 $\pm$ 0.8 & 53.3 $\pm$ 0.4 \\
Otters+HAT$^1$  & 89.3 $\pm$ 0.5 & \textbf{89.0} $\pm$ \textbf{0.6} & 85.3 $\pm$ 0.6 & 76.5 $\pm$ 0.4 & 61.0 $\pm$ 0.8 \\
Otters+HAT$^2$  & 87.4 $\pm$ 0.5 & 87.2 $\pm$ 0.7 & \textbf{85.9} $\pm$ \textbf{0.7} & \textbf{85.2} $\pm$ \textbf{0.7} & \textbf{80.8} $\pm$ \textbf{0.3} \\
\midrule
\multicolumn{6}{c}{\textbf{ Parameter } $\boldsymbol{\beta}$ \textbf{ Experiments}} \\
\cmidrule(lr){1-6}
$\beta$ & \textbf{0.01} & \textbf{0.02} & \textbf{0.03} & \textbf{0.04} & \textbf{0.05} \\
\midrule
Otters & \textbf{90.2} $\pm$ \textbf{0.1} & \textbf{89.5} $\pm$ \textbf{0.1} & 87.5 $\pm$ 1.8 & 79.2 $\pm$ 3.4 & 72.5 $\pm$ 1.4 \\
Otters+HAT$^1$& 89.6 $\pm$ 0.1 & 89.2 $\pm$ 0.3 & \textbf{89.1} $\pm$ \textbf{0.4} & \textbf{87.7} $\pm$ \textbf{1.2} & 80.6 $\pm$ 2.8 \\
Otters+HAT$^2$ & 87.8 $\pm$ 0.7 & 88.1 $\pm$ 0.4 & 87.4 $\pm$ 1.0 & 85.5 $\pm$ 1.1 & \textbf{83.8} $\pm$ \textbf{0.5} \\
\midrule
\multicolumn{6}{c}{\textbf{Parameters } $\boldsymbol{\tau}$ \textbf{ Experiments}} \\
\cmidrule(lr){1-6}
Otters \textbf{0.10} & \textbf{0.20} & \textbf{0.30} & \textbf{0.40} & \textbf{0.50} \\
\midrule
$\tau$ + GN & \textbf{90.3} $\pm$ \textbf{0.1} & \textbf{90.1} $\pm$ \textbf{0.5} & 87.0 $\pm$ 1.7 & 75.0 $\pm$ 4.1 & 67.5 $\pm$ 4.8 \\
Otters+HAT$^1$ & 89.6 $\pm$ 0.5 & 89.3 $\pm$ 1.2 & \textbf{88.2} $\pm$ \textbf{0.7} & \textbf{88.2} $\pm$ \textbf{1.0} & 76.1 $\pm$ 2.8 \\
Otters+HAT$^2$  & 87.8 $\pm$ 0.6 & 87.7 $\pm$ 0.8 & 87.1 $\pm$ 0.6 & 87.2 $\pm$ 0.3 & \textbf{83.0} $\pm$ \textbf{1.4} \\
\bottomrule
\end{tabular}%
}
\end{table*}

\section{Conclusion}
This paper introduces Otters, a new paradigm for energy-efficient neuromorphic computing that challenges this digital-centric approach. Through a hardware-software co-design, we repurpose the natural signal decay of a custom-fabricated optoelectronic synapse, transforming this physical phenomenon  into a computational method. This allows us to eliminate the costly decay function evaluation steps inherent in traditional TTFS, fusing computation and memory into the physical process. To deploy this paradigm in complex architectures like the Transformer, we developed a  QNN-to-SNN conversion algorithm that circumvents the challenges of direct SNN training. The Otters model achieves state-of-the-art accuracy across seven GLUE benchmark datasets among SNNs, while simultaneously delivering a 1.77$\times$ improvement in energy efficiency  over previous leading spiking models. By directly harnessing fundamental device physics for computation, this work demonstrates a new path to a more energy-efficient neuromorphic computing design.

\clearpage


\bibliography{iclr2026_conference}
\bibliographystyle{iclr2026_conference}

\clearpage

\appendix
\section{Appendix}
\subsection{Fabricated In$_2$O$_3$ TFTs }
\label{sec:Fabricated }
To prepare the indium oxide thin-film transistors (In$_2$O$_3$ TFTs), indium nitrate was first dissolved in a mixed solvent of 2-methoxyethanol (2-ME), acetylacetone (AcAc), and ammonium hydroxide (NH$_3$·H$_2$O) to form a precursor solution, which was stirred overnight to ensure complete dissolution and coordination. Subsequently, gate electrodes were fabricated on a silicon substrate coated with a SiO$_2$ insulating layer, followed by sequential deposition of 8~nm chromium and 50~nm gold via electron-beam evaporation. A 30~nm-thick Al$_2$O$_3$ dielectric layer was then uniformly deposited over the substrate using atomic layer deposition (ALD). The In$_2$O$_3$ precursor solution was spin-coated onto the dielectric surface, after which the channel regions were defined through standard photolithography, and the unprotected areas were removed by hydrochloric acid wet etching. The films were annealed in air at 300~$^\circ$C for 1~hour to enhance crystallinity and improve film quality. Portions of the Al$_2$O$_3$ layer were subsequently etched to expose selected regions of the gate electrodes. Finally, source and drain electrodes, along with interconnects, were patterned and metallized with an additional 8~nm chromium and 50~nm gold layer via electron-beam evaporation. The indium oxide thin-film transistor was characterized under a gate bias of 0~V and a drain bias of 5~mV. Upon 405~nm laser illumination, oxygen vacancies in the channel layer were photoionized, generating free electrons and thereby enhancing the channel conductivity. 



\begin{figure}[h!]
    \centering
    \begin{subfigure}[b]{0.49\textwidth}
        \centering
    \includegraphics[width=0.68\textwidth]{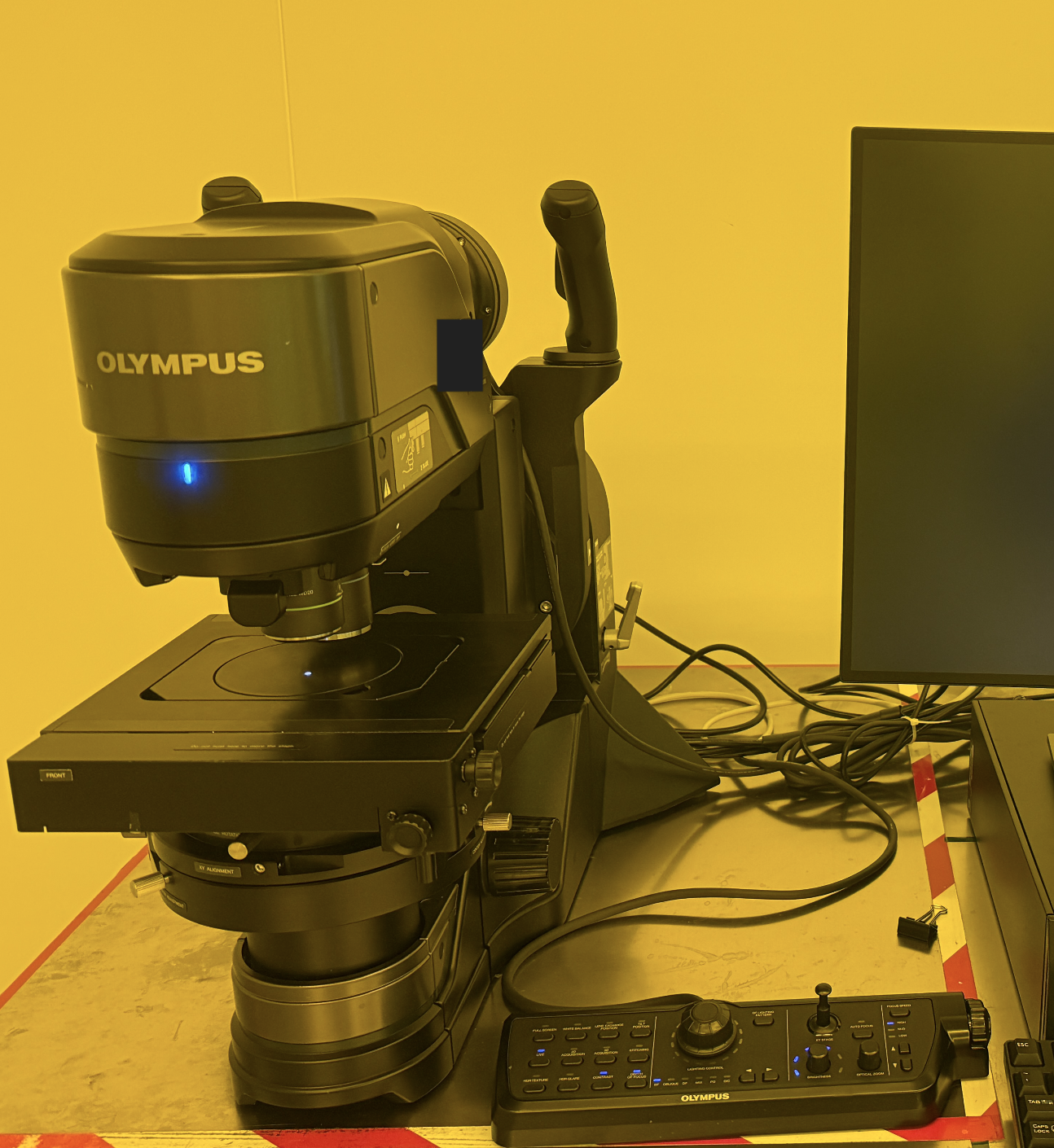}
        \caption{Optical microscope}
    \end{subfigure}
    \hfill
    \begin{subfigure}[b]{0.49\textwidth}
        \centering
        \includegraphics[width=1.09\textwidth]{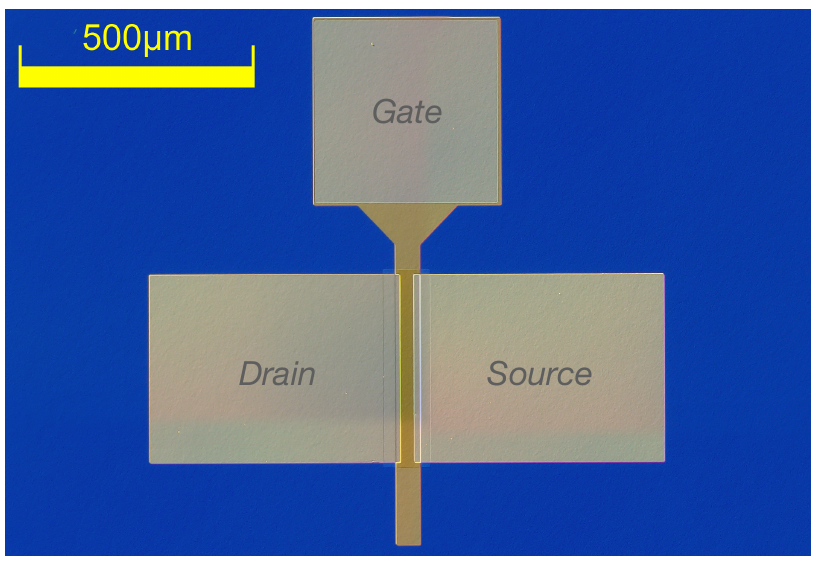}
        \caption{Otters under optical microscope}
        \label{fig:optical}
    \end{subfigure}
    \caption{Details design of Otters}
\end{figure}


\subsection{Proof for Proposition 1}
\label{sec: proof}
\subsubsection{Integration Phase Equivalence}
The foundation of the proof lies in the relationship between the QNN's discrete value and the SNN's spike time. From Section~\ref{sec:qnn_to_snn}, we know that the QNN output activation from the previous layer is $x_{q,i}^{l-1} = \alpha^{l-1} \cdot q_i^{l-1}$, where $q_i^{l-1}$ is the integer value:
\[
    q_i^{l-1} = \text{Clip}\left(\left\lfloor \frac{a_i^{l-1}}{\alpha^{l-1}} \right\rfloor, 0, T\right)
\]
We define the TTFS encoding scheme such that this integer value $q_i^{l-1}$ is represented by a single spike from neuron $i$ at the discrete time step $k$:
\[
    k = T - q_i^{l-1}
\]
This encoding adheres to the TTFS principle: a larger integer value $q_i^{l-1}$ results in a smaller time step $k$, signifying an earlier spike. A value of 0 corresponds to no spike within the active window, and the maximum value $T$ corresponds to a spike at $k=0$.

When a presynaptic neuron $i$ fires at time step $k$, its contribution to the postsynaptic potential of neuron $j$ is given by the \textit{Otters} PSP function, $\epsilon'(\cdot)$. Using the conditions specified in Proposition 1, the normalized value produced by the decay function $O(t)$ at time step $t_k$ is:
\[
    O(t_k) = \frac{T-k}{T}
\]
Substituting the encoding relationship from Step 1 ($k = T - q_i^{l-1}$):
\[
    O(t_k) = \frac{T - (T - q_i^{l-1})}{T} = \frac{q_i^{l-1}}{T}
\]
Thus we find that the normalized output of the physical decay process at the spike time $t_k$ is directly proportional to the integer value $q_i^{l-1}$ it is meant to encode.

The full PSP contribution from the synapse connecting $i$ to $j$ is the product of this normalized value and the scaling factor $\gamma_{ij}^l$. Using the definition of $\gamma$ from Proposition 1:
\[
    \gamma_{ij}^l = w_{ij}^l \cdot \alpha^{l-1} \cdot T
\]
The PSP is therefore:

\begin{align*}
    \epsilon'(w_{ij}^l, t_k) &= \gamma_{ij}^l \cdot O(t_k) \\
    &= \left(w_{ij}^l \cdot \alpha^{l-1} \cdot T\right) \cdot \left(\frac{q_i^{l-1}}{T}\right) \\
    &= w_{ij}^l \cdot (\alpha^{l-1} \cdot q_i^{l-1}) \\
    \intertext{Recognizing that $x_{q,i}^{l-1} = \alpha^{l-1} \cdot q_i^{l-1}$, we find:}
    \epsilon'(w_{ij}^l, t_k) &= w_{ij}^l \cdot x_{q,i}^{l-1}
\end{align*}
This shows that the contribution of a single spike in the SNN is exactly equal to the weighted input term in the QNN.

The final membrane potential $V_j^l$ is the sum of all such PSPs from incoming spikes, plus the bias term:
\[
    V_j^l = \sum_i \epsilon'(w_{ij}^l, t_{\text{spike},i}^{l-1}) + b_j^l = \sum_i (w_{ij}^l \cdot x_{q,i}^{l-1}) + b_j^l
\]
By comparing this with the definition of the QNN pre-activation from Section 1.2,
\[
    a_j^l = \sum_i w_{ij}^l x_{q,i}^{l-1} + b_j^l
\]
we arrive at the desired equality:
\[
    V_j^l = a_j^l
\]

\subsubsection{Firing phase equivalence}
In this section, we prove that the integer value encoded by the SNN's output spike time, $t_{\text{spike},j}^l$, is equal to the integer value of the QNN's output, $q_j^l$. That is, if $t_{\text{spike},j}^l$ corresponds to time step $k_{\text{fire}}$, we have:
\[
    T - k_{\text{fire}} = q_j^l = \text{Clip}\left(\left\lfloor \frac{a_j^l}{\alpha^l} \right\rfloor, 0, T\right)
\]

According to the SNN model definition, the neuron fires at the earliest discrete time step $k$ for which its potential $V_j^l$ meets or exceeds the threshold $\theta^l(t_k)$.
\[
    V_j^l \ge \theta^l(t_k)
\]
Substituting the result from Part I ($V_j^l = a_j^l$) and the definition of the time-varying threshold from Proposition 1 ($\theta^l(t_k) = \alpha^l \cdot (T-k)$), the firing condition becomes:
\[
    a_j^l \ge \alpha^l \cdot (T-k)
\]

Assuming $\alpha^l > 0$, we can rearrange the inequality to solve for the term $(T-k)$, which represents the integer value that would be encoded by a spike at time step $k$:
\[
    \frac{a_j^l}{\alpha^l} \ge T-k
\]

The threshold $\theta^l(t_k) = \alpha^l (T-k)$ is a monotonically decreasing function of the time step $k$. For a fixed membrane potential $a_j^l$, this means that if the firing condition is met for a certain time step $k^*$, it will also be met for all subsequent time steps $k > k^*$. The TTFS firing rule dictates that the neuron fires at the first time step that satisfies the condition. This corresponds to finding the smallest integer $k$ that satisfies the inequality (largest \(T-k\) ). By the definition of the floor function, the integer value encoded by the output spike, which is defined by our encoding scheme as $q_{\text{out},j}^l = T - k_{\text{fire}}$, is:
\[
    q_{j}^l = \left\lfloor \frac{a_j^l}{\alpha^l} \right\rfloor
\]

The derivation above assumes the result of the floor function falls within the valid range of encodable integers. We now analyze the boundary conditions imposed by the finite simulation window.

\begin{itemize}
    \item Upper Bound (Clipping at T) If the pre-activation $a_j^l$ is very large such that $\lfloor a_j^l / \alpha^l \rfloor > T$, the condition $a_j^l / \alpha^l \ge T-k$ will be satisfied for all $k \in T$. The neuron will fire at the earliest possible time step, which is $k=0$. The value encoded by a spike at $k=0$ is $T-0=T$. This naturally implements the upper bound of the clipping function, mapping any integer value greater than $T$ to $T$.

    \item Lower Bound (Clipping at 0). If the pre-activation $a_j^l$ is such that $\lfloor a_j^l / \alpha^l \rfloor < 0$ , then the term $a_j^l / \alpha^l$ is negative. Thus, the firing condition $a_j^l / \alpha^l \ge T-k$ can never be satisfied. The neuron will not fire within the time window. The absence of a spike is interpreted as encoding the integer value 0. This naturally implements the lower bound of the clipping function.
\end{itemize}

Combining these cases, the integer value encoded by the SNN's firing mechanism, $q'_{\text{out},j}$, is:
\[
    q'_{\text{out},j} = \text{Clip}\left(\left\lfloor \frac{a_j^l}{\alpha^l} \right\rfloor, 0, T\right)
\]
This is identical to the definition of the QNN's integer output, $q_j^l$.

\subsection{Energy Analysis}
\label{sec: energy_qnn}
\subsubsection{Energy comparison with related works}
We compare our model, Otter, against two classes of baselines: standard Transformers and Spiking Transformers.

For transformer baselines, the energy consumption of a Full BERT (FP32) is dominated by expensive 32-bit multiply-accumulate (MAC) operations and data movement. Another primary baseline is a Quantized BERT (INT4), which shares the same architectural settings as Otter, utilizing 4-bit activations and a 1-bit Key-Value (KV) cache.

\textbf{Full BERT (FP32)}
\[
E_{FC} = B \cdot S \cdot C_o \cdot (\gamma \cdot C_i \cdot (E_{\text{MAC}}+E^{\text{Read}}_{\text{weight}}+32 \cdot E^{\text{sparse}}_{\text{move}})+C_i\cdot E_{\text{leakage}}+2E_{\text{clamp}}+E^{\text{write}}_{kv})
\]
\[
E_{FC-score} = B \cdot h \cdot S^2 \cdot (d_k \cdot \gamma \cdot (E^{\text{Read}}_{\text{kv}}+E_{\text{MAC}}+32\cdot E^{\text{sparse}}_{\text{move}})+d_k\cdot E_{\text{leakage}}+2E_{\text{clamp}})
\]
\textbf{Quantized BERT }
\[
E_{FC_{q}} = B \cdot S \cdot C_o \cdot (\gamma \cdot C_i \cdot (E_{\text{MAC}}+E^{\text{Read}}_{\text{weight}}+\log_2(T+1)E^{\text{sparse}}_{\text{move}})+C_i \cdot E_{\text{leakage}}+2E_{\text{clamp}}+E^{\text{write}}_{kv})
\]
\[
E_{FC_q-score} =  B \cdot h \cdot S^2 \cdot (d_k \cdot \gamma \cdot (E^{\text{Read}}_{\text{kv}}+E_{\text{MAC}}+\log_2(T+1) \cdot E^{\text{sparse}}_{\text{move}})+d_k \cdot E_{\text{leakage}}+2E_{\text{clamp}})
\]

For Spiking Neural Network Baselines such as Sorbet, SpikingBERT, and SpikingLM, energy cost is a function of the spike rate ($s_r$ ) and the number of timesteps ($T$), with expenses driven by accumulation and memory access. To create a fair comparison, we pick the most optimized setting of these models (Sorbet, SpikingBERT, and SpikingLM) which match Otter's performance, resulting in required spike rates of 13\% for Sorbet, 25\% for SpikingBERT, and 33\% for SpikingLM. The number of timesteps (T) for each model is set to the value from its original paper: 16, 16, and 4, respectively.

\textbf{Typical SNNs}
\[
E_{\text{SNN-FC}} = B \cdot S \cdot C_o \cdot (C_i \cdot s_r \cdot T \cdot (E_{\text{ACC}}+E^{\text{Read}}_{\text{weight}}+E^{\text{sparse}}_{\text{move}})+ C_i\cdot T \cdot E_{\text{leakage}} +T \cdot (E_{\text{CMP}}+s \cdot E_{\text{SUB}})+E^{\text{Write}}_{\text{kv}})
\]
\[
E_{\text{SNN-score}} =  B \cdot h \cdot S^2 \cdot (d_k \cdot s_r \cdot T \cdot (E^{\text{Read}}_{\text{kv}}+E_{\text{ACC}}+E^{\text{sparse}}_{\text{move}})+d_k\cdot T \cdot E_{\text{leakage}}+ T \cdot (E_{\text{CMP}}+s \cdot E_{\text{SUB}}))
\]

\textbf{Otters}

\[
\begin{split}
    E_{\text{Opto-FC}} = \underbrace{B \cdot S \cdot C_o}_{\text{Total Outputs}} \cdot \bigg( 
    &\underbrace{C_i \cdot T \cdot \big(s_r \cdot (E_{\text{ACC}} + E^{\text{Read}}_{\text{Analog}} + E^{\text{sparse}}_{\text{move}}) }_{\text{Spike Processing}}  \\
    & + E_{\text{leakage}}\big) + \underbrace{T \cdot (E_{\text{CMP}}+E^{\text{Read}}_{\text{threshold}})}_{\text{Thresholding}} + \underbrace{E^{\text{Write}}_{\text{binarykv}}}_{\text{K/V Write}} \bigg)
\end{split}
\]

\[
\begin{split}
    E_{\text{Opto-score}} = \underbrace{B \cdot h \cdot S^2}_{\text{Total Scores}} \cdot \bigg( 
    &\underbrace{d_k \cdot T \cdot \big(s_r \cdot (E_{\text{ACC}} + E^{\text{Read}}_{\text{Analog}} + E^{\text{sparse}}_{\text{move}} + E^{\text{Read}}_{\text{binarykv}}) }_{\text{Spike Processing}} \\+ E_{\text{leakage}} \big)
    & + \underbrace{T \cdot (E_{\text{CMP}}+E^{\text{Read}}_{\text{threshold}})}_{\text{Thresholding}} \bigg)
\end{split}
\]

Key differences between Otters and other typical SNNs include replacing digital weight reads with lower-energy \text{analog reads} from the TFT ($E^{\text{Read}}_{\text{Analog}}$) and, for the QKV calculation, using an energy-efficient \text{binary KV read} ($E^{\text{Read}}_{\text{binarykv}}$). The energy model is configured for a BERT-base architecture with a batch size ($B$) of 64, a sequence length ($S$) of 128, and input/output channel dimensions ($C_i, C_o$) of 768. The model features 12 attention heads ($h$), with a per-head dimension ($d_k$) of 64. Energy costs are derived from established models. For FP32 operations, we assume that a multiply-accumulate (MAC) consumes 4.6~pJ and a clamp operation consumes 0.9~pJ which is from ~\citep{horowitz20141} cause our platform doesn't support FP calculations. For our INT4 models, we differentiate between a 4-4-16bits MAC (0.0848~pJ) and a 1-4-16bits MAC (0.0663~pJ). The costs for 4-16-16bits, 2-16-16bits and 1-16-16bits ACC are 0.0502~pJ, 0.0477~pJ and 0.0429~pJ, respectively. 
SNN-specific operations, such as comparison and subtraction, are each modeled at 0.0502~pJ. The total energy for an analog read operation is 0.0246~pJ (0.00875pJ for power the TFT, 1.33e-6pJ for sampling, 0.0053pJ for ADC including the amplifier and 0.010505 for the 4bits LUT). 
The model accounts for a static leakage energy \(E_{\text{leakage}}\) of 0.002~pJ per cycle, a weight activation (read/write) cost of 0.0985~pJ/bit, and a sparse data movement cost of 0.18~pJ per bit. 
All compute, data movement, and memory access costs using energy measurements from a commercial 22nm process. 

\subsubsection{Ablation study of using Otters and traditional TTFS methods}
Traditional TTFS requires digital encoding, additional MAC operations (\(E_{\text{encoding}}{+}E_{\text{MAC}}\)), and extra weight accesses. With \(T{=}15\) using the simplest \(T{-}t\) encoding (4--4--4-bit ACC, \(0.0163\,\mathrm{pJ}\)), traditional TTFS attention block consumes \(5.12\,\mathrm{mJ}\), \(26.1\%\) more energy than 1bit Otters.

\subsection{Forward process of Otters}
\label{sec: forward}
The forward process is shown in Algorithm~\ref{alg:otters_neuron}: 
\begin{algorithm}[h!]
\caption{Otters Neuron Forward Pass (TTFS)}
\label{alg:otters_neuron}
\begin{algorithmic}[1]
\STATE \textbf{Inputs:} Presynaptic spikes $\{s_i^{l-1}(t)\}_{i=1}^{N_{in}}$; weights $\{\gamma_{ij}^l\}_{i=1}^{N_{in}}$; bias $b_j^l$; threshold function $\theta^l(t)$; total time steps $T$.
\STATE \textbf{Output:} Postsynaptic spike train $\{s_j^{l}(t)\}_{t=1}^{T}$.
\STATE $V_j \gets b_j^l$ \COMMENT{Initialize membrane potential with bias}
\STATE $s_j^{l}[1:T] \gets 0$ \COMMENT{Initialize output spike train to zeros}
\STATE \texttt{has\_fired} $\gets \textbf{false}$
\FOR{$t = 1, 2, \dots, T$}
    \FOR{$i = 1, 2, \dots, N_{in}$}
        \IF{$s_i^{l-1}(t) = 1$}
            \STATE $v_i \gets \gamma_{ij}^l \cdot O(t)$ \COMMENT{Compute PSP using the physical decay function $O(t)$}
            \STATE $V_j \gets V_j + v_i$ \COMMENT{Accumulate potential}
        \ENDIF
    \ENDFOR
    \IF{$V_j \ge \theta^l(t)$ \textbf{and not} \texttt{has\_fired}}
        \STATE $s_j^{l}(t) \gets 1$ \COMMENT{Fire a spike at the current time step}
        \STATE \texttt{has\_fired} $\gets \textbf{true}$
        \STATE \textbf{break} \COMMENT{TTFS constraint: emit at most one spike and stop}
    \ENDIF
\ENDFOR
\STATE \textbf{return} $s_j^{l}$
\end{algorithmic}
\end{algorithm}

\subsection{QNN training algorithm}
\label{sec: training algo}
\begin{table}[h!]
\centering
\caption{Task-specific training hyperparameters.}
\label{tab:hyperparams}
\begin{tabular}{lccc}
\toprule
\textbf{Task} & \textbf{Max. Seq. Length} & \textbf{Batch Size} & \textbf{Learning Rate} \\
\midrule
MNLI  & 128 & 80  & $2 \times 10^{-4}$ \\
MRPC  & 128 & 80  & $5 \times 10^{-5}$ \\
SST-2 & 64  & 80 & $1 \times 10^{-4}$ \\
STS-B & 128 & 80  & $5 \times 10^{-5}$ \\
QQP   & 128 & 80  & $2 \times 10^{-5}$ \\
QNLI  & 128 & 80  & $2 \times 10^{-4}$ \\
RTE   & 128 & 80  & $5 \times 10^{-5}$ \\
\bottomrule
\end{tabular}
\end{table}

To mitigate the accuracy degradation from quantization, we employ a knowledge distillation (KD) framework for our quantization-aware training~\citep{liu2022bit,tangsorbet}. This approach transfers inductive biases from a pre-trained, full-precision teacher model to the quantized student network. Our training objective is a hybrid loss function that aligns both the output distributions and the intermediate representations of the student with those of the teacher. The total distillation loss $L_{KD}$ is a weighted combination of two components: a logits distillation loss $L_{\text{logits}}$ and a representation distillation loss $L_{\text{reps}}$:
\begin{equation}
    L_{KD} = L_{\text{logits}} + \lambda L_{\text{reps}}
\end{equation}
where $\lambda$ is a hyperparameter that balances the two terms.

The first component, \textbf{$L_{\text{logits}}$}, is the Kullback-Leibler (KL) divergence between the teacher's soft target distribution $p$ and the student's output distribution $q$. This loss encourages the student to learn the teacher's predictions and its understanding of inter-class similarities.
\begin{equation}
    L_{\text{logits}} = \text{KL}(p || q)
\end{equation}
The second component, \textbf{$L_{\text{reps}}$}, minimizes the Mean Squared Error (MSE) between the intermediate feature representations from corresponding transformer blocks of the teacher ($r_i^t$) and the student ($r_i^s$). This forces the student to mimic the teacher's internal representation structure.
\begin{equation}
    L_{\text{reps}} = \sum_i \|r_i^t - r_i^s\|_2^2
\end{equation}

The student model is trained end-to-end by minimizing $L_{KD}$ using the Adam optimizer, while the teacher model's weights remain frozen. We use task-specific hyperparameters for training, which are detailed in Table~\ref{tab:hyperparams}. All models are trained on A100 GPUs.

\subsection{Evaluation Benchmark}
\label{sec: datasets}
We evaluated our model, Otters, on seven datasets from the GLUE benchmark:

\begin{itemize}
    \item \text{MNLI (Multi-Genre Natural Language Inference):} A large-scale, crowdsourced collection of sentence pairs annotated for textual entailment across multiple genres.
    
    \item \text{QQP (Quora Question Pairs):} A paraphrase identification task to determine if two questions from Quora are semantically equivalent.
    
    \item \text{QNLI (Question-Answering NLI):} A natural language inference task converted from the Stanford Question Answering Dataset (SQuAD), where the goal is natural language inference
    
    \item \text{SST-2 (Stanford Sentiment Treebank):} A single-sentence classification task for sentiment analysis (positive or negative) on movie reviews.
    
    \item \text{STS-B (Semantic Textual Similarity Benchmark):} A regression task to predict the degree of similarity (on a 1--5 scale) between sentence pairs drawn from news headlines, video titles, and image captions.
    
    \item \text{RTE (Recognizing Textual Entailment):} A compilation of data from several textual entailment challenges, using text from news articles and Wikipedia.
    
    \item \text{MRPC (Microsoft Research Paraphrase Corpus):} A paraphrase detection task using sentence pairs from online news sources. The dataset is imbalanced, with 68\% of pairs being paraphrases.
\end{itemize}

\subsection{The Use of Large Language Models}
We use LLMs to aid or polish writing.
\end{document}